\documentclass{llncs}
\usepackage{graphicx}

\usepackage{tikz}
\usepackage{comment}
\usepackage{amsmath,amssymb} 
\usepackage{color}
\usepackage{epsfig}
\usepackage{amsmath}
\usepackage{amssymb}
\usepackage{booktabs}
\usepackage[ruled]{algorithm2e}
\usepackage{multirow}
\usepackage{pdfpages}

\usepackage{algorithmic}

\begin{document}

\makeatletter

\def\blfootnote{\gdef\@thefnmark{}\@footnotetext}

\makeatother

\pagestyle{headings}
\mainmatter
\def\ECCVSubNumber{2489}  

\title{Efficient Ensemble Model Generation for Uncertainty Estimation with Bayesian Approximation in Segmentation} 

\author{Hong Joo Lee\textsuperscript{*}\textsuperscript{1}, Seong Tae Kim\textsuperscript{*}\textsuperscript{2}, Hakmin Lee\textsuperscript{1}, Nassir Navab\textsuperscript{2,3}, \\ and Yong Man Ro\textsuperscript{1}}
\institute{\textsuperscript{1} Image and Video Systems Lab, KAIST, South Korea \\ \email{\{dlghdwn008, zpqlam12, ymro\}@kaist.ac.kr}\\
\textsuperscript{2} Computer Aided Medical Procedures, Technical University of Munich, Germany \\ \email{seongtae.kim@tum.de, navab@cs.tum.edu}\\
\textsuperscript{3} Computer Aided Medical Procedures, Johns Hopkins University, USA}

\maketitle
\blfootnote{*Both authors equally contributed to this paper.}
\begin{abstract}
Recent studies have shown that ensemble approaches could not only improve accuracy and but also estimate model uncertainty in deep learning. However, it requires a large number of parameters according to the increase of ensemble models for better prediction and uncertainty estimation. To address this issue, a generic and efficient segmentation framework to construct ensemble segmentation models is devised in this paper. In the proposed method, ensemble models can be efficiently generated by using the stochastic layer selection method. The ensemble models are trained to estimate uncertainty through Bayesian approximation. Moreover, to overcome its limitation from uncertain instances, we devise a new pixel-wise uncertainty loss, which improves the predictive performance. To evaluate our method, comprehensive and comparative experiments have been conducted on two datasets. Experimental results show that the proposed method could provide useful uncertainty information by Bayesian approximation with the efficient ensemble model generation and improve the predictive performance.
\keywords{Uncertainty estimation, ensemble model, stochastic layer selection, segmentation}
\end{abstract}

\section{Introduction}
Semantic segmentation is widely used in real-world applications such as road scene analysis and medical diagnosis. With the development of deep learning technology, considerable accurate segmentation methods have been reported \cite{chen2017deeplab,chen2018encoder,yuan2019object,badrinarayanan2017segnet,ronneberger2015u,long2015fully,yu2015multi}. Although these methods have achieved high performance in public benchmark datasets, the deep networks sometimes predict overly confident results although the decision is not correct. Such overconfident predictions lead to critical problems in safety-critical applications such as medical diagnosis or autonomous driving. For example, in the autonomous driving system, the network may segment a person as a tree with high probability. For better behavior decisions, it is desirable to know the model uncertainty and reject uncertain predictions. Also, in the medical applications, if a model can indicate uncertain decisions whether it is a lesion or not, the physicians would widely adopt the model in clinical work-flows. Therefore, it is useful to know the uncertainty of segmentation models in real-world applications. 

To address this issue, several studies have been reported to estimate the model uncertainty \cite{gal2016dropout,kendall2017bayesian,rupprecht2017learning,kendall2017uncertainties,bhattacharyya2018long,feng2018towards,kendall2018multi,nair2020exploring,roy2019bayesian,postels2019sampling}. There are two approaches to estimate the model uncertainty: 1) Bayesian approach \cite{gal2016dropout,kendall2017uncertainties,bhattacharyya2018long,kendall2017bayesian}, 2) ensemble approach \cite{lakshminarayanan2017simple}. The Bayesian approaches are widely used tools to estimate the uncertainty of models \cite{welling2011bayesian}. The main purpose of the Bayesian approaches is to approximate the posterior distribution of the network weight parameters. However, the Bayesian approaches are not easy to handle directly in the deep neural networks (DNNs) since it requires a large computational cost. To handle the uncertainty in the DNNs, \cite{gal2016dropout} approximates Bayesian inference by using dropout \cite{srivastava2014dropout}. 
This stochastic regularization inference is referred to as Monte Carlo dropout (MC-dropout). Inspired by this work, \cite{kendall2017bayesian} proposed Bayesian approximation for semantic segmentation by applying MC-dropout to convolution layers. 
The model uncertainty is estimated by calculating the variance of each pixel. \cite{nair2020exploring} also reported the quality control method by using the uncertainty estimation with MC-dropout in medical segmentation tasks. Bayesian approaches are presented based on the assumption that the dropout rates are well-tuned based on the training data to estimate model uncertainty \cite{lakshminarayanan2017simple}. 

Recent studies have shown that the ensemble approaches are effective to improve accuracy and estimate model uncertainty \cite{ashukha2020pitfalls,valdenegro2019deep,fort2019deep}. In ensemble studies\cite{lakshminarayanan2017simple,gustafsson2019evaluating}, $M$ different models which have different weight parameters could be constructed and the mean of the predictions is used for the final prediction. The variance could be used as the uncertainty of the model. Lakshminarayanan et al. \cite{lakshminarayanan2017simple} have proposed a simple and scalable non-bayesian approach to estimate model uncertainty with ensemble models for regression and classification tasks. The posterior of the model’s parameters is estimated via random changes during the training. As a result, they improve the predictive uncertainty for detecting out-of-distribution inputs. Although this method has achieved successes in estimating uncertainty with a simple implementation, it requires a large number of parameters according to the increase of ensemble models for better prediction. To estimate the uncertainty with a large number of samples, the number of network parameters linearly grows.

Most of these studies have focused on estimating the uncertainty of the model at the inference. A few studies have investigated the use of uncertainty to improve the model in training, e.g., robust learning under noisy data \cite{kendall2017uncertainties}, multitask learning \cite{kendall2018multi}, and confidence calibration \cite{seo2019learning}. However, it has not been thoroughly investigated to use the uncertainty for model improvement by guiding the model gradually overcoming its limitation. Note that a problem of learning from uncertainty is not well defined in general because it is an unsupervised problem where each model could have different ground-truth for uncertain and certain samples. It should be updated according to the time during the training. This is similar to the human whose uncertainty assessments tend to be personal and vary over time according to the experience. Therefore, an uncertainty estimator should be operated in an unsupervised manner and model specific. 

In this paper, we focus on tackling the problem of parameter increase in the ensemble model for segmentation tasks. For this purpose, a generic and efficient segmentation framework to construct ensemble segmentation models is devised. We define a network model set which consists of \textit{M} submodels with \textit{L} layers. Each submodel has the same encoder-decoder architecture with different weights. Ensemble models can be generated by stochastically selecting layers from the submodels in the model set. To train the model set that can predict accurately and estimate model uncertainty, we propose a novel two-stage training method by leveraging uncertainty for the segmentation tasks. In the first step, various networks are constructed by stochastic layer selection method, and the network model set is optimized to be able to predict reliable results over various ensemble models. By selecting each layer according to the Bernoulli distribution, the networks are trained to estimate uncertainty through Bayesian approximation. 
In the second step, we devise a pixel-wise uncertainty loss to train the model in the segmentation task. In the segmentation task, we can make good use of pixel-wise uncertainty estimation that indicate uncertain pixel. Instead of learning the model to perform on every pixel at once, the pixel-wise uncertainty loss guides the model to gradually overcome its limitation from uncertain pixels, which improves the segmentation performance. At the inference time, we can ideally construct ${{M}^{L}}$ ensemble models with \textit{M} submodels through the stochastic layer selection method. The proposed model samples the prediction with randomly selected layers by Monte Carlo (MC) sampling, which can estimate the predictive uncertainty. Our contributions are summarized as follows:
\begin{itemize}
    \item We propose a novel segmentation framework to efficiently generate ensemble models by using the stochastic layer selection method. We can effectively construct ${{M}^{L}}$ ensemble models with \textit{M} submodels and estimate model uncertainty through Bayesian approximation.
    \item We propose a novel pixel-wise uncertainty loss, which encourages the segmentation model to focus on uncertain regions. By leveraging the uncertainty, the segmentation model could gradually learn from uncertain pixels which is easier than hard negative samples in training.
    \item By comprehensive and comparative experiments, we verify the effectiveness of the proposed method for both the performance improvement and the segmentation quality control.
\end{itemize}

\section{Related Work}
\textbf{Uncertainty Estimation.}
Uncertainty estimation is an essential issue for bringing deep learning technology to real-world applications such as autonomous driving and medical diagnosis. To address this issue, considerable research efforts have been devoted to uncertainty modeling and estimation in deep neural networks. Bayesian approach \cite{srivastava2014dropout,graves2011practical,blundell2015weight,hernandez2015probabilistic,gal2016dropout} is a representative method for uncertainty estimation with a mathematical framework. Bayesian neural networks model parameter uncertainty explicitly by approximately learning the posterior distribution of the parameters, which induce uncertainty over neural network activations and predictions. 
Gal and Ghahramani \cite{gal2016dropout} proposed a simple way to estimate uncertainty by an MC-dropout sampling. The MC-dropout sampling simply approximates Bayesian neural networks by using dropout \cite{srivastava2014dropout} at the inference time. Inspired by this works, the MC-dropout sampling method has been widely used in various applications including medical segmentation \cite{nair2020exploring,roy2019bayesian}, autonomous driving \cite{bhattacharyya2018long,feng2018towards}, multi-task learning \cite{kendall2018multi}, and active learning \cite{gal2017deep,kirsch2019batchbald}. Kendall and Gal have extended this Bayesian approximation to computer vision tasks and differentiate the epistemic uncertainty with aleatoric uncertainty \cite{kendall2017uncertainties}. 
From these studies, MC-dropout sampling approaches are presented based on the assumption that the dropout rates are well-tuned based on the training data to estimate model uncertainty.

\textbf{Ensemble Modeling.}
Deep ensembles have shown to be effective to improve accuracy \cite{lakshminarayanan2017simple,fort2019deep}, estimate uncertainty \cite{osband2016deep,lakshminarayanan2017simple,rupprecht2017learning}, and increase robustness to adversarial examples \cite{pang2019improving,tramer2017ensemble,lakshminarayanan2017simple}. Lakshminarayanan et al. \cite{lakshminarayanan2017simple} have proposed ensemble modeling as a non-Bayesian solution to quantify predictive uncertainty. Ilg et al. \cite{rupprecht2017learning} utilized boot-strapped ensembles to predict uncertainty in optical flow prediction. Ovadia et al. \cite{snoek2019can} and Gustafsson et al. \cite{gustafsson2019evaluating} observed that ensemble models tend to outperform than Bayesian neural networks in terms of accuracy and uncertainty estimation, particularly under dataset shift. Compared with Bayesian modeling which assumes that the true model lies within the hypothesis class of the prior, and performs soft model selection to find the single best model within the hypothesis class \cite{minka2000bayesian}, ensembles perform the model combination. In other words, ensembles combine the models to obtain a more powerful model and it can be expected to be better when the true model does not lie within the hypothesis class\cite{lakshminarayanan2017simple}. Although these ensemble approaches show superior results in terms of uncertainty estimation and improve accuracy, it requires a large number of parameters according to the increase of ensemble models for better prediction. To estimate the uncertainty with a large number of samples, the number of network parameters linearly grows.

\textbf{Segmentation.}
Segmentation is widely used in real-world applications such as road scene analysis and medical diagnosis. Due to its importance, many accurate algorithms have been reported \cite{zhao2017pyramid,chen2017deeplab,zhu2019improving,yuan2019object}. 
Most of these methods focus on improving segmentation accuracy.
Some works have handled uncertainty estimation in segmentation task \cite{nair2020exploring,roy2019bayesian,kendall2017bayesian,huang2018efficient} because of its importance. \cite{kendall2017bayesian} has proposed Bayesian approximation for semantic segmentation. To estimate the posterior distribution of pixel class labels, they applied Monte Carlo sampling with dropout. Huang et al. \cite{huang2018efficient} proposed fast uncertainty estimation methods for video segmentation. They utilized the temporal information based on the properties of video continuity. They showed their effectiveness by using Tiramusi \cite{jegou2017one} and Bayesian SegNet \cite{kendall2017bayesian} on CamVid dataset \cite{brostow2009semantic}.

\begin{figure*}[t]
	\centering
	\includegraphics[width=0.9\textwidth]{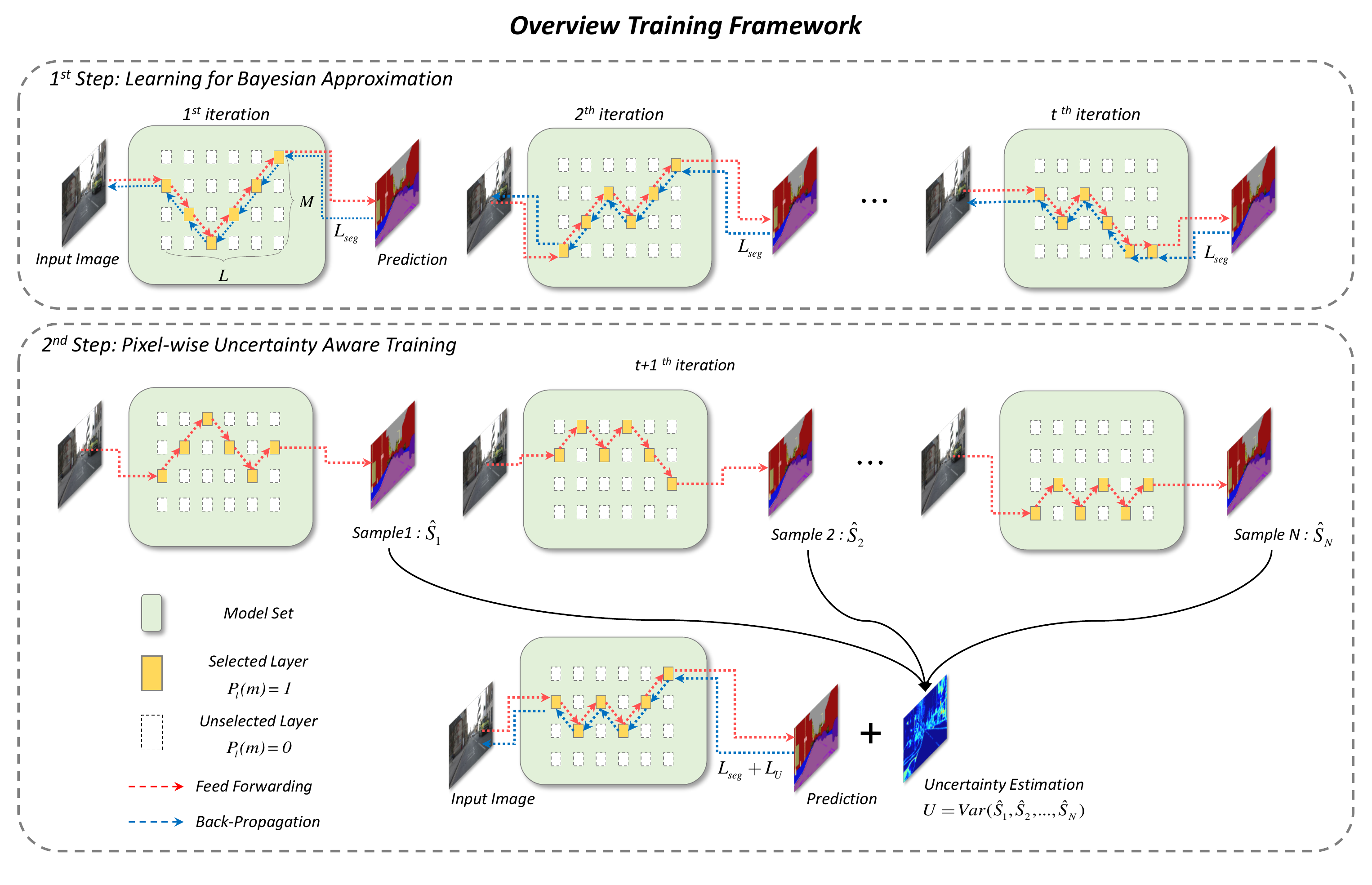}
	\vspace{-0.5cm}
	\caption{Overview of the proposed training framework. It consists of learning for Bayesian approximation and pixel-wise uncertainty aware training.}
	\vspace{-0.2cm}
	\label{fig1}
\end{figure*} 
\section{Proposed Method}

In this section, we describe our proposed stochastic layer selection method and learning strategy. Figure \ref{fig1} shows an overview of the proposed training framework. As shown in the Figure \ref{fig1}, a model set consist of $M$ submodels with $L$ layers. The model set is trained with two steps in our method. In the first step, the model set is constructed by randomly selected layers for each iteration. The model set is optimized by a segmentation loss function to be able to reliably predict the output from arbitrary models, which consist of random combinations of layers. In this step the model set is learned to predict segmentation map and estimate uncertainty. Then, in the second step, the model set is optimized by two loss functions. One is the segmentation loss function that is used in the first step and the other one is a pixel-wise uncertainty loss that encourages the network to learn from the uncertain part of the predicted segmentation map. These loss values are back-propagated to selected layers and update the corresponding weight parameters. For the testing, we sample $N$ segmentation maps with randomly selected layers. We use the mean of samples as a segmentation prediction and use the pixel variance as a predictive uncertainty. 
\subsection{Stochastic Layer Selection}
In this section, we describe how to design the model set and how to select layers stochastically. The proposed model set consists of $M$ (where $M\ge 2$) parallel submodels and each submodel has $L$ layers. Then, we can sample a model with random variable ${P}_{l}(m)$ as follow

\begin{equation}\label{eq1}
\begin{aligned}
{{P}_{l}}(m)=\left\{ \begin{matrix}
   1,\ \text{if}\ Layer_{m}^{l}\ \text{is}\ \text{selected,}  \\
   0,\ \text{otherwise}.\,\,\,\,\,\,\,\,\,\,\,\,\,\,\,\,\,\,\,\,\,\,\,\,\,\,\,\   \\
\end{matrix} \right.
\end{aligned} 
\end{equation}

In order to ensure that each layer is selected from one submodel, we constrain the ${{P}_{l}}(m)$ as follow

\begin{equation}\label{eq2}
\begin{aligned}
\sum\limits_{m=1}^{M}{{{P}_{l}}(m)=1},
\end{aligned}
\end{equation}
\begin{equation}\label{eq3}
\begin{aligned}
{P}_{l}\sim categorical({{x}_{1}},{{x}_{2}},...,{{x}_{M}};{{\mu }_{1}},{{\mu }_{2}},...,{{\mu }_{M}}),
\end{aligned}
\end{equation}
where ${{P}_{l}}$ is a $1\times M$ dimension vector of categorical random variables with probability ${{\mu }_{m}}$ of being 1. Then, the total weight ($W$) and random selecting vector of the model set can be defined as follow
\begin{equation}
    \begin{aligned}
        W=
\begin{Bmatrix}
{w}_{1,1} & \dots  & {w}_{1,L}\\ 
\vdots & \ddots & \vdots \\ 
{w}_{M,1} & \dots &{w}_{M,L}
\end{Bmatrix},
    \end{aligned}
\end{equation}
\begin{equation}
    \begin{aligned}
    P=\begin{Bmatrix}
{P}_{1}\\ 
\vdots\\ 
{P}_{L}
\end{Bmatrix},
\end{aligned}
\end{equation}
where ${w}_{m,l}$ denotes the weight of $Layer_{m}^{l}$. Then, we sample a model which has same structure of sub model with $P_{l}(m)$. The weights of sampled model can be represented as follow
\begin{equation}
    {w}^{Sample} = diag(P\times W).
\end{equation}

In contrast to dropout method that apply masking the perceptrons, our proposed method applies categorical random variables to each layer. We can construct ${{M}^{L}}$ ensemble models only with $M$ submodels which have same structure while have different weights. Furthermore, since the distribution of categorical random variables follows Bernoulli distribution, the model set can estimate the posterior distribution over the model weight parameters by Bayesian approximation. Therefore, our model can leverage both the effect of ensemble model that improve predictive performance \cite{dietterich2000ensemble,lakshminarayanan2017simple}, and Bayesian approximation.
\subsection{Uncertainty Measure using Stochastic Layer Selection}
By the stochastic layer selection method, we can construct a large number of ensemble models within the limited number of parameters. It is well known fact that ensemble model can predict uncertainty by approximating the posterior with an ensemble of sample distributions \cite{efron1994introduction,osband2016deep,lakshminarayanan2017simple}. We leverage the ensemble learning to perform our probabilistic inference. For given trained model set, we sample $N$ models with the random layer selection as in Eq.\ref{eq1} and predict $N$ segmentation prediction $\mathbf{S}=\{{{\hat{S}}_{1}},{{\hat{S}}_{2}},...,{{\hat{S}}_{N}}\}$. To estimate uncertainty, we take a variance of predicted samples as $U=var\{{{\hat{S}}_{1}},{{\hat{S}}_{2}},...,{{\hat{S}}_{N}}\}$. 
\subsection{Learning with Stochastic Layer Selection}
As shown in Figure \ref{fig1}, our proposed stochastic layer selection learning method consists of two steps. In the first step, we train model set to be able to predict reliable segmentation map and estimate uncertainty by Bayesian approximation. In the second step, we utilize predictive uncertainty to encourage the model to focus on uncertain predictions and gradually overcome its limitation.
\subsubsection{Learning for Bayesian Approximation}
In the first step, we train the model set to be able to predict uncertainty. To this end, we find the posterior distribution over the model weight parameters $W$, with the given training data $X$, and training label $Y$. The posterior distribution can be defined as $p(w|X,Y)$. We can estimate the posterior by variational approximation. This approximation can be formulated as follow

\begin{equation}\label{eq4}
\begin{aligned}
{\mathcal{L}_{\operatorname{var}}}=-\sum\limits_{i=1}^{K}{\int\limits_{w}{{{q}_{\theta }}(w)\log p({{y}_{i}}|{{x}_{i}},w)dw}}+KL({{q}_{\theta }}(w)||p(w)),
\end{aligned}
\end{equation}
where ${{q}_{\theta }}(w)$ denotes approximate posterior and $KL({{q}_{\theta }}(w)||p(w))$ denotes Kullback-Leibler (KL) divergence with true posterior. $K$ denotes the number of example in dataset. However, this form of distribution is intractable, since we have to integrate over the all data. Therefore, we approximate the posterior distribution by Monte Carlo method. This approximation can be written as 

\begin{equation}\label{eq5}
\begin{aligned}
{\mathcal{L}_{\operatorname{var}}}=-\frac{1}{BN}\sum\limits_{i=1}^{B}{{}}\sum\limits_{j=1}^{N}{\log p({{y}_{i}}|{{x}_{i}},{{{\hat{w}}}_{i,j}})}+KL({{q}_{\theta }}(w)||p(w)),
\end{aligned}
\end{equation}
where $B$ and $N$ denote the size of a batch and the number of samples, respectively. ${{\hat{w}}_{i,j}}\sim {{q}_{\theta }}(w)$ denotes a sample from the approximate distribution. Note that randomized weights that have Bernoulli distribution can approximate model of the Gaussian process \cite{gal2016bayesian}. Then, it can be approximated by minimizing the KL divergence. 

Since we select the layers along with categorical random distribution ($P_{l}$) that follows Bernoulli distribution, we can approximate posterior by replacing $\hat{w}_{i,j}\leftarrow {w}^{Sample}$ in Eq. \ref{eq5}. Therefore, we approximate our posterior by minimizing KL divergence. Since minimizing the KL divergence term has the same effect of minimizing the cross entropy loss \cite{gal2016bayesian} , we optimize the model by reducing the cross entropy loss with our stochastic layer selection inference. Therefore, in the first step, we train the model with cross entropy loss. It enables the model set reliably predict the segmentation map and estimate uncertainty. This step continues until the validation loss is converged.
\subsubsection{Pixel-wise Uncertainty Aware Training}
After the first training step is finished, in the second step, we improve the segmentation prediction by pixel-wise uncertainty loss, as illustrated in Figure \ref{fig1}. To this end, during the training, we generate pixel-wise uncertainty map $U_{i}=var\{{{\hat{S}}_{1}},{{\hat{S}}_{2}},...,{{\hat{S}}_{N}}\}$, where ${\hat{S}}_{n}$ denotes the sample prediction with randomly seleced layers. $U_{i}$ means the pixel-wise variations between $N$ samples. With the pixel-wise uncertainty map, the pixel-wise uncertainty loss function can be defined as follows

\begin{equation}\label{eq6}
\begin{aligned}
{\mathcal{L}_{U}}=-\sum\limits_{i=1}{({{U}_{i}}\otimes {{Y}_{i}})\log ({{U}_{i}}\otimes {{{\hat{Y}}}_{i}})},
\end{aligned}
\end{equation}

\begin{equation}\label{eq7}
\begin{aligned}
{{\hat{Y}}_{i}}=f({{X}_{i}},{{P}_{1}},{{P}_{2}},...,{{P}_{L}}),
\end{aligned}
\end{equation}
where $i$ denotes the index of training image, and $f(\cdot )$ is a function that maps the given input image $X$ to segmentation map with selected layers. $\otimes$ denotes an element-wise multiplication. The re-weighting of uncertain areas with large weights allows the model to improve on these samples. Therefore, the pixel-wise uncertainty loss encourages the model gradually to improve its limitation. Finally, in the second training step, a total loss can be written as follows,

\begin{equation}\label{eq8}
\begin{aligned}
{\mathcal{L}_{total}}=\underbrace{-\sum\limits_{i=1}{{{Y}_{i}}\log ({{{\hat{Y}}}_{i}})}}_\text{Cross Entropy}-\underbrace{\lambda \sum\limits_{i=1}{({{U}_{i}}\otimes {{Y}_{i}})\log ({{U}_{i}}\otimes {{{\hat{Y}}}_{i}})}}_\text{Pixel-wise Uncertainty},  
\end{aligned}
\end{equation}
where $\lambda$ denotes a balancing hyper-parameter. The algorithms for the training and the test are described in Appendix A.
%

\section{Experiment}
To verify the effectiveness of the proposed method, we evaluate our method on two possible safe-critical applications. Firstly, for the autonomous driving application, we use CamVid\cite{brostow2009semantic} dataset which is widely used for road scene understanding and uncertainty estimation. For the medical diagnosis, we use Transvaginal Ultrasound (TVUS) dataset which is an in-house dataset for the experiment. 
\subsection{Experiment Setting}
\subsubsection{CamVid Dataset}
The dataset is a publically available semantic segmentation dataset for road scene understanding, which has applications for autonomous driving. The dataset is also widely used to analyze the predictive uncertainty \cite{kendall2017bayesian,huang2018efficient,lakshminarayanan2017simple,postels2019sampling}. It consists of 367 training and 233 testing images of day and dusk scenes. Following the experimental protocol in \cite{badrinarayanan2017segnet}, we use 11 classes such as road, building, cars, etc. To avoid overfitting, we conduct data augmentation (vertical flip, random crop, resize, and random rotation).

For the model set design, we use the FC-DenseNet 103 \cite{jegou2017one} as a submodel. Then, we construct a model set with two submodels (\textit{M}=2). We define the layers as 'Dense Block' unit for our model set. Therefore, in our experiments we set \textit{M}=2 and \textit{L}=11. This model set is annotated as DenseNet-2 in the following experiment table and figure. We optimize the model set by using RMS-Prop \cite{tieleman2012lecture} with a learning rate of 0.001 and exponential decay of 0.995 after each epoch. To get the mean prediction and predictive uncertainty, we sample 50 times \textit{N}=50.

\subsubsection{TVUS Dataset}
The private dataset is collected from three hospitals. The purpose of this dataset is to segment endometrium regions on Transvaginal Ultrasound (TVUS) image. The dataset can be used to diagnose infertility by considering the shape and size of the endometrium. All the images are annotated by experienced gynecologists. To minimize the annotation inconsistency between gynecologists, we define the 4 gynecological points (cavity tip, internal os, and two thickest points between the two basal layers) \cite{machtinger2005transvaginal,gonen1989endometrial,park2019endometrium} with the gynecologists. Then, they tried to annotate the GT mask while considering the point definition. After that, they cross-checked the annotations of each other and eliminated inconsistent annotations. As a result, we could build a well-annotated database with 3,372 images and corresponding endometrium segmentation maps. The full size of the image is 256$\times$320. To avoid overfitting, we also conduct data augmentation (vertical \& horizontal flip, random rotation). For the evaluation, we conduct 5 folds cross-validation. 

For the model set design, we use an U-Net architecture \cite{ronneberger2015u} which is widely used for medical segmentation. Then, we construct a model set with two submodels ($M=2$). We duplicate the every convolution layer in U-Net. We train the model from scratch without any preprocessing. Therefore, in our experiments we set $M=2$ and $L=18$. This model set is annotated as U-Net-2. We optimize our model set using ADAM optimizer \cite{kingma2014adam} with a fixed learning rate 0.0001. To get the mean prediction and predictive uncertainty, we sample 50 times ($N=50$).
\subsection{Quantitative Evaluation}
\begin{table}[t]
	\centering
	\caption{Comparison of Mean Intersection over Union (mIoU) and the number of network parameter on CamVid dataset.}
	\vspace{-0.2cm}
	\scalebox{0.8}{
		\begin{tabular}{l|c|c}
			\toprule
			\multicolumn{1}{c|}{\multirow{2}{*}{\textbf{Method}}} & \multirow{2}{*}{\textbf{mIoU}} &\textbf{Number of}\\&&\textbf{Parameter (M)}\\
			\midrule
			SegNet \cite{badrinarayanan2017segnet} & 46.4&29.5 \\
			FCN-8s \cite{long2015fully} & 57.0&134.5 \\
			DeepLab-LFOV \cite{chen2017deeplab} & 61.6&37.3 \\
			DFANet \cite{li2019dfanet} & 64.7&- \\
			Dilation8 \cite{yu2015multi} & 65.3&- \\
			Dilation8 + FSO \cite{kundu2016feature} & 66.1&140.8 \\
			DenseNet \cite{jegou2017one}(Our Implementation)& 66.9(66.4)&9.4\\
			G-FRNet \cite{amirul2017gated} & 68.0&29.5 \\
			Bi-Senet \cite{yu2018bisenet} & 68.7&49.0 \\
			\hline
			\multicolumn{2}{l}{\textit{Uncertainty-Based Method}} \\
			\midrule
			Bayesian SegNet \cite{kendall2017bayesian} & 63.1 &29.5\\
			\textbf{Bayesian SegNet-2 (ours)}  & 64.5 &59.0\\
			\textbf{Bayesian SegNet-2 + Pixel-wise}& \multicolumn{1}{c|}{\multirow{2}{*}{65.3}}&\multicolumn{1}{c}{\multirow{2}{*}{59.0}} \\
			 \textbf{Uncertainty Loss (ours)} &  & \\
            \midrule
			Kendall et al. \cite{kendall2017uncertainties} & 67.5&9.4\\
			\textbf{DenseNet-2 (ours)} & 67.4 &18.8\\
			\textbf{DenseNet-2 + Pixel-wise}& \multicolumn{1}{c|}{\multirow{2}{*}{68.2}}&\multicolumn{1}{c}{\multirow{2}{*}{18.8}} \\
			 \textbf{Uncertainty Loss (ours)} &  & \\
			\bottomrule
	\end{tabular}}
    \vspace{-0.3cm}
	\label{camvid}

\end{table}
\begin{table}[t]
	\centering
	\caption{Dice and Jaccard coefficient comparison of our method and other methods on TVUS dataset. The fourth column denotes the number of each network parameter.}
	\vspace{-0.3cm}
	\resizebox{0.8\linewidth}{!}{
		\begin{tabular}{l|c|c|c}
			\toprule
			\multicolumn{1}{c|}{\multirow{2}{*}{\textbf{Method}}}          & \multicolumn{1}{c|}{\multirow{2}{*}{\textbf{Dice Coefficient}}} & \multicolumn{1}{c|}{\multirow{2}{*}{\textbf{Jaccard Coefficient}}}&\textbf{Number of}\\&&&\textbf{Parameter (M)} \\ \toprule
			U-Net \cite{ronneberger2015u} & 82.30 & 70.38&31.0 \\
			FCN-8s \cite{long2015fully} & 81.19 & 69.12&134.5 \\
			Dilation8 \cite{yu2015multi} & 82.40 & 70.36&140.8 \\
			Endometrium SegNet \cite{park2019endometrium} & 82.67 & 70.46&- \\ \midrule
			\multicolumn{1}{l}{\textit{Uncertainty-Based Method}} \\ \midrule
			U-Net + MC-dropout \cite{nair2020exploring} & 82.00 & 68.92&31.0 \\
			\textbf{U-Net-2 (ours)}  & 82.72 & 70.09 &62.0 \\
			\textbf{U-Net-2 + Pixel-wise} &  \multicolumn{1}{c|}{\multirow{2}{*}{\textbf{83.51}}}&\multicolumn{1}{c|}{\multirow{2}{*}{\textbf{71.40}}}&\multicolumn{1}{c}{\multirow{2}{*}{62.0}} \\ 
			\textbf{Uncertainty Loss (ours)} &  & \\ \bottomrule
	\end{tabular}}
\end{table}
To demonstrate the advantage of proposed method, we compare with other methods on two different datasets. Table 1 shows the mean intersection over union (mIoU) score on Camvid dataset. The DenseNet-2 is trained by our stochastic layer selection method without pixel-wise uncertainty loss. The DenseNet-2 + Pixel-wise Uncertainty Loss is trained by our stochastic layer selection method with pixel-wise uncertainty loss. In the case of DenseNet-2 + Pixel-wise Uncertainty Loss, the mIoU score achieves 68.2\%. It has been improved to 1.3\% compared to original DenseNet. Also, compared with other semantic segmentation networks, our method shows comparable performance. Furthermore, we compare the proposed method with recent uncertainty-based semantic segmentation methods. Our method shows better performance than other uncertainty-based methods. We observe that, compare to DenseNet-2, DenseNet-2+Pixelwise Uncertainty Loss shows better performance. This result can be interpreted in a way that our pixel-wise uncertainty loss improves the network effectively by gradually reducing the uncertain region during training. Also, to verify applicability of the proposed method, we applied the proposed method to Bayesian SegNet \cite{kendall2017bayesian}. The results show that our method also achieves better performance than Bayesian SegNet, which indicates that our method could be applicable to other segmentation model.

For further evaluation of our method in medical application, we conduct more experiments on a challenging TVUS dataset. Table 2 shows the comparison with the conventional segmentation networks (FCN-8s and Dilation8) and an Endometrium SegNet according to Dice Coefficient Score (DCS) and Jaccard Coefficient Score (JCS). In the case of U-Net-2 + Pixel-wise Uncertainty Loss, we achieve 83.51\% in DCS and 71.40\% in JCS respectively. It has improved to 1.21\% compared to the original U-Net. To prove that the performance improvement is statistical significance, we conduct a paired t-test. The paired t-test provides the statistical evaluation and qualification procedure of the segmentation network. The performance improvement is statistically significant ($p$\textless{}0.01 by paired t-test). Our method is also higher than U-Net+MC-dropout which samples 50 times with dropout. The proposed U-Net + Pixel-wise Uncertainty Loss shows 1.51\% higher performance than U-Net+MC-dropout \cite{nair2020exploring}. Furthermore, it shows superior performance compared to Endometrium SegNet \cite{park2019endometrium}. Since our method encourages the model to gradually learn the uncertain part by pixel-wise uncertainty loss, it shows better performance than other methods.

\subsection{Quantitative Comparison with Ensemble Model}
\begin{table}[t]
	\centering
	\caption{Comparison with ensemble model and our proposed method.}
	\vspace{-0.3cm}
    \resizebox{0.8\linewidth}{!}{
    \begin{tabular}{c|c|c|c|c}
        \hline
        \multirow{2}{*}{\textbf{\# of Network}} & \multicolumn{2}{c|}{\textbf{CamVid (DenseNet)}} & \multicolumn{2}{c}{\textbf{TVUS (U-Net)}} \\ \cline{2-5} 

        & \textbf{mIOU} & \textbf{\# of Parameters (M)} & \textbf{Dice Coefficient} & \textbf{\# of Parameters (M)} \\ 
        \midrule
        Ensemble (1)  & 66.4              & 9.4                  & 82.30        & 31.0  \\                   
        Ensemble (2)  & 66.8              & 18.8                 & 82.51        & 62.0  \\                 
        Ensemble (3)  & 67.0              & 28.2                 & 83.12        & 93.1  \\                  
        Ensemble (4)  & 67.1              & 37.6                 & 83.20        & 124.1 \\                 
        Ensemble (5)  & 67.2              & 47.0                 & 83.38        & 155.1 \\
        \midrule
        \textbf{Ours}          & \textbf{68.2}              & 18.8                 & \textbf{83.51}        & 62.0 \\
        \bottomrule
    \end{tabular}}
    \vspace{-0.2cm}
\end{table}
One of the our main contributions is that our model efficiently generates ensemble models. Therefore, in this section, we verify the effectiveness of our method by comparing with the vanilla ensemble models in terms of the number of parameters and the performance. To this end, we trained five different networks (DenseNet, U-Net) separately and estimate mean prediction and uncertainty with 5 samples. For each training setting, we use the same training hyper-parameters, e.g., learning rate, training iteration, optimizer, etc. The initial weights are set differently for each network. Table 3 shows the experiment results according to the number of sub-network in the ensemble model. As shown in Table 3, in the case of CamVid dataset with DensNet, our method shows 1.4\% higher mIoU than Ensemble (2) which has the same number of parameters. Also our method shows 0.13\% higher mIoU than Ensemble (5) which has a large number of parameters than our method. In the case of TVUS dataset with U-Net, our method shows 1\% higher mIoU than Ensemble (2) which has the same number of parameter. Also our method shows 0.2\% higher mIoU than Ensemble (5) which has a large number of parameters than our method. These results show that the proposed method effectively generates the ensemble models.
\subsection{Quantitative Evaluation with Uncertainty-based Pixel Rejection}

In this section, we demonstrate that our method effectively predicts the uncertain regions. To verify this, we conduct uncertainty-based pixel rejection experiment. For the pixel rejection experiment, we normalize the estimated uncertainty map from 0 to 1. Then, by sorting the uncertainty in descending order, we remove the pixels that the models indicate high uncertainty. We evaluate the performance on the remaining pixels in testing images.

In the CamVid dataset, we reject the uncertain pixels from 0\% to 20\% of total testing image pixels with an interval of 2.5\%. Then, we measure the mIoU with remaining pixels. Figure 2 (a) shows the pixel rejection results on CamVid dataset. For the DenseNet + MC-dropout and DenseNet + Pixel-wise Uncertainty Loss, we estimate uncertainty with 50 samples. The DenseNet Ensemble (5) is a simple ensemble model with five different DenseNet network. We train the networks separately with random weight initialization. Then, we estimate the uncertainty with variance of each network prediction. In the case of random rejection, we randomly remove the pixel. As shown in Figure 2 (a), removing the pixel randomly does not help to improve the performance. In the case of DenseNet Ensemble (5), by removing the uncertain pixel, the mIoU is increased. It shows better performance than MC-dropout method at the beginning. However, since the number of samples is limited, DensNet Ensemble (5) could not estimate uncertainty well than MC-dropout method. Note that the number of parameter increases to increase the number of sample in the vanilla ensemble model. The MC-dropout method shows better performance than ensemble model from the moment 10\% of uncertain pixels are removed. In the case of our method, as the uncertain pixels are removed, the mIoU score increased. Also, it shows better performance than MC-dropout method. 

In the TVUS dataset, we reject the uncertain pixels from 0\% to 5\% of total testing image pixels with an interval of 0.5\%. Figure 2 (b) shows the pixel rejection results on TVUS dataset. For the U-Net + MC-dropout and U-Net + Pixel-wise Uncertainty Loss, we estimate with 50 samples. In the case of U-Net Ensemble (5), we trained five different U-net separately and estimate uncertainty with 5 samples. As shown in Figure 2 (b), the random rejection does not help to improve the performance. Also, in the case of the ensemble networks, the result shows same tendency as shown in the previous experiment. It could not indicate uncertain pixels well than MC-dropout method since its limitation of the number of samples. In the case of our method, as the uncertain pixels are removed, the dice coefficient score is increased. Also, it shows better performance than MC-dropout method. Note that although our method utilizes the uncertainty to improve the model performance in training, the uncertainty at the inference time is still meaningful to indicate the confidence of the model’s prediction. The proposed method can guarantee the performance by rejecting to operate on samples where it is likely to fail.

\begin{figure*}[t]
	\centering
	\includegraphics[width=0.7\textwidth]{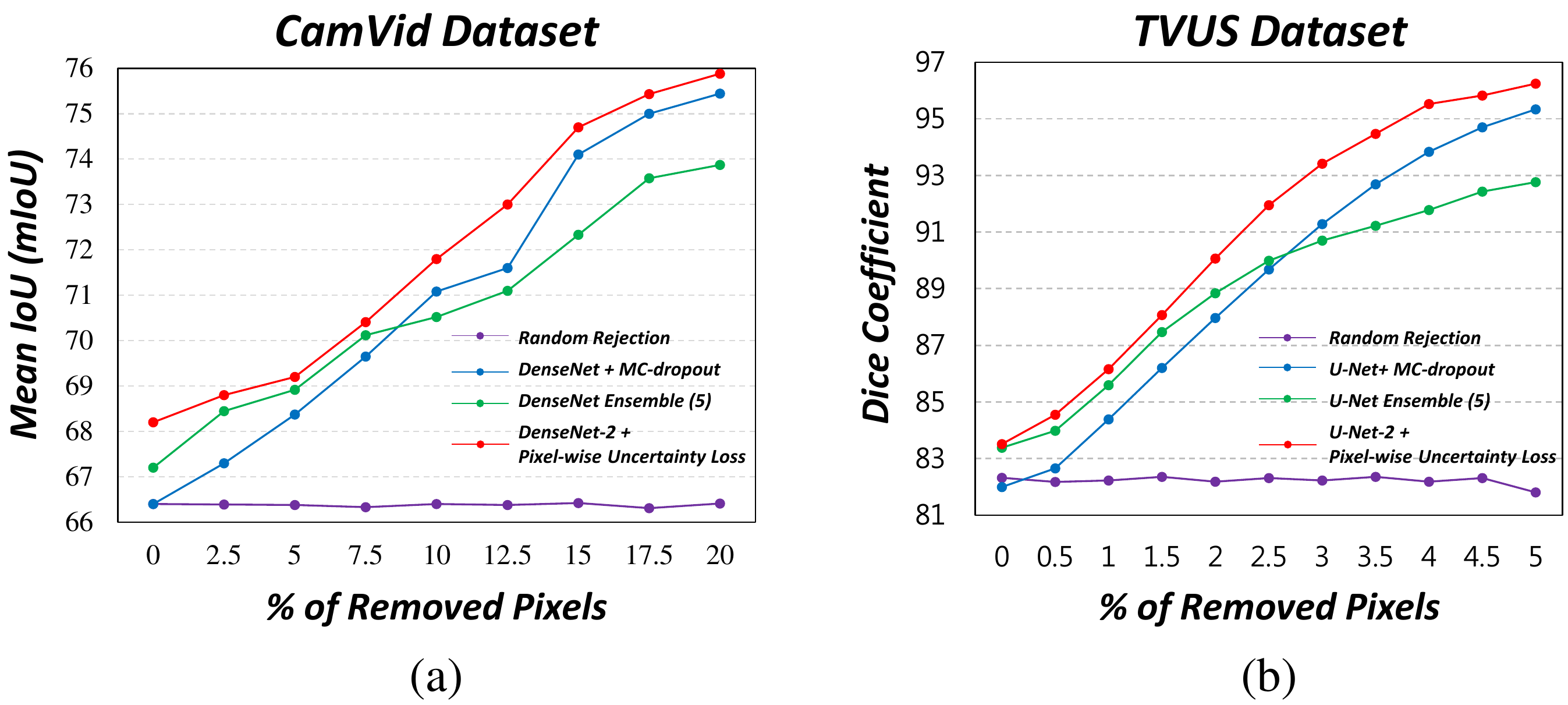}
    \vspace{-0.2cm}
	\caption{(a) Mean IOU according to pixel-wise rejection on CamVid dataset. (b) Dice coefﬁcient according to pixel-wise rejection on TVUS dataset.}
	\label{fig2}
\end{figure*} 
\subsection{Qualitative Results}
\begin{figure*}[t]
\begin{minipage}[b]{0.48\linewidth}
	\centering
	\includegraphics[width=0.84\textwidth]{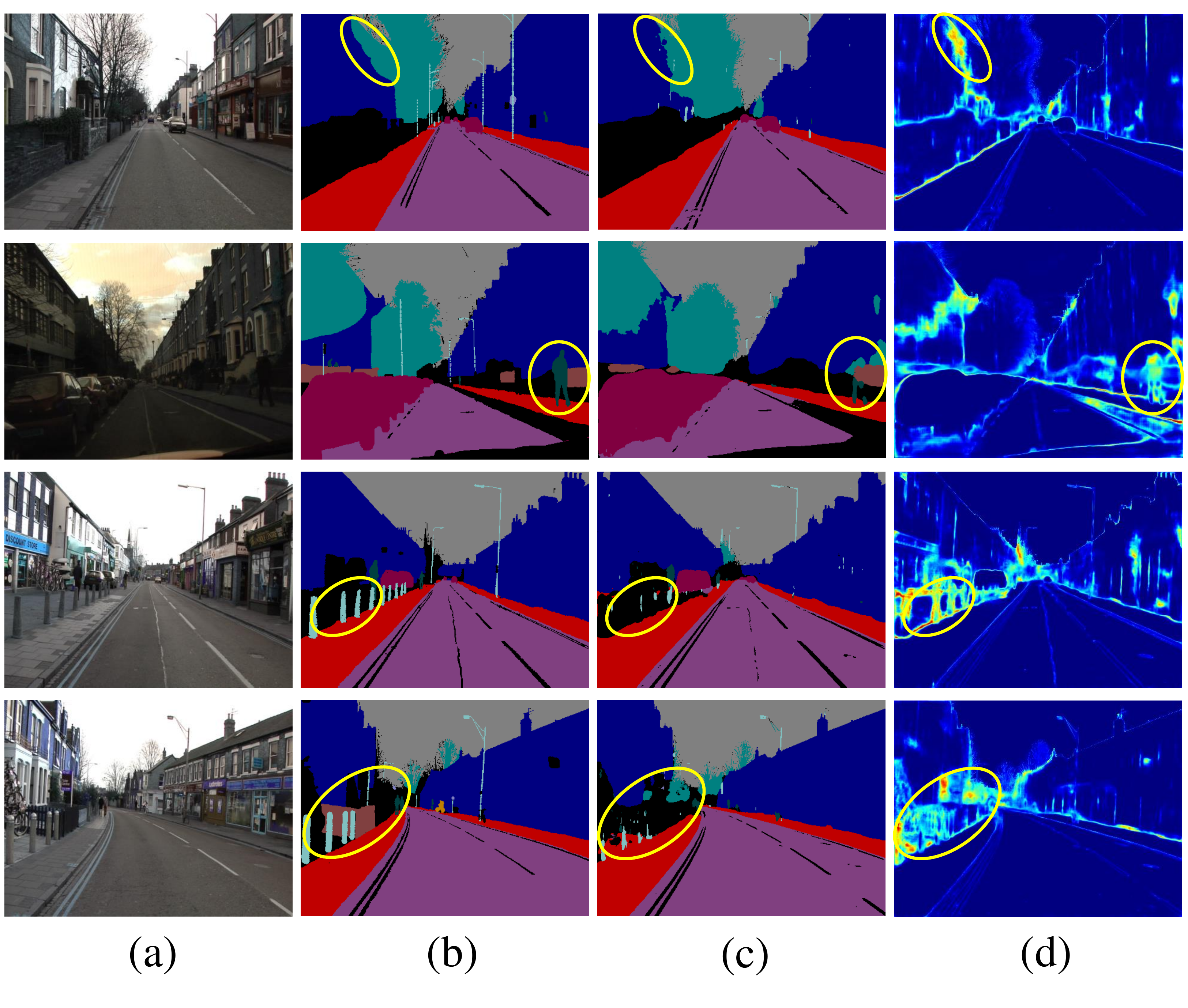}
    
	\caption{Qualitative results of segmentation and uncertainty estimation in our method on CamVid dataset. (a) is input images, (b) is ground truth maps, (c) is our mean prediction results with 50 samples, and (d) is our uncertainty map estimation results. The yellow circle denotes incorrect segmentation regions.}
	\vspace{-0.1cm}
	\label{fig3}
	\end{minipage}
	\hspace{0cm}
	\begin{minipage}[b]{0.48\linewidth}
	\centering
    
	\includegraphics[width=1\textwidth]{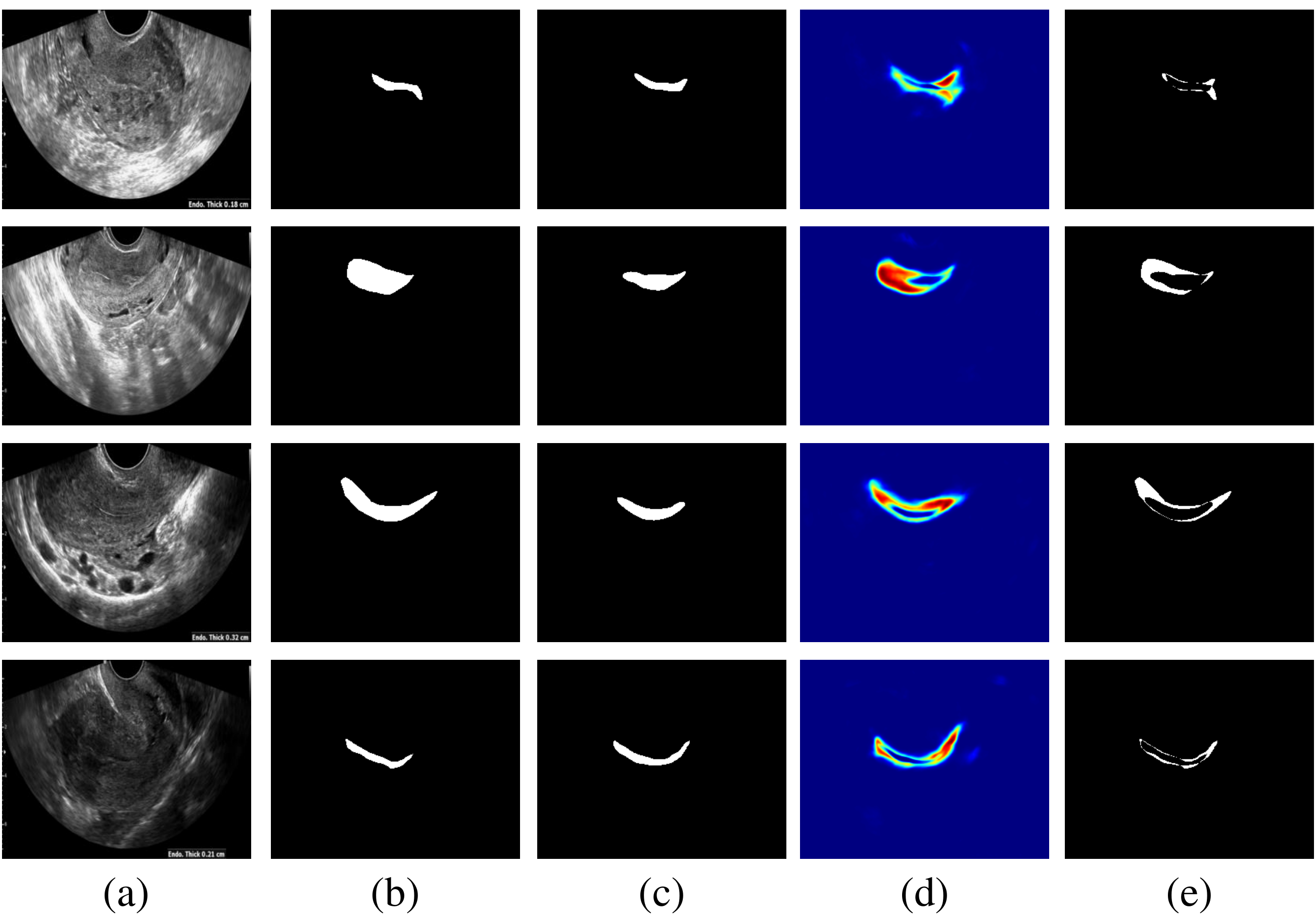}
    
	\caption{Qualitative results of segmentation and uncertainty estimation in our method on TVUS dataset. (a) is input images, (b) ground truth maps, (c) is mean prediction results with 50 samples, (d) uncertainty map by the proposed method, (e) is the difference between ground truth and segmentation results.}
	\vspace{-0.1cm}
	\label{fig4}
	\end{minipage}
\end{figure*} 

In this section, we visually observe the segmentation results and uncertainty estimation results. Figure 3 shows the predicted segmentation map and corresponding uncertainty map on CamVid dataset. 
As shown in the figure, although the network predicts incorrect segmentation results, the network estimate that the region is uncertain. In the case of the first and second rows, the network estimates uncertain regions where the object is hard to be observed because of the surrounding background and illumination. In the case of the third and fourth row, the network estimates uncertain regions where the unseen classes (black color in ground truth map) appeared. This allows the user to make comprehensive decisions by utilizing the uncertainty as well as predicted results.

For further evaluation, we visually observe the segmentation results and uncertainty estimation results on TVUS dataset. Figure 4 shows the predicted segmentation map and corresponding uncertainty map. 
As shown in the figure, the uncertain region and mis-segmentation region are similar. These results indicate that although the network predicts incorrectly, the user can make comprehensive decisions by utilizing the uncertainty.

\subsection{Effect of number of sample}

In this section, we analyze the effect of the number of sample in terms of mean prediction performance. Figure 5 (a) shows the mIoU score on CamVid dataset. Figure 5 (b) shows Dice Coefficient Score on TVUS dataset. Both results show that as the number of increase, the performance also increases. Then, our proposed method shows better performance with a few number of sample. In the Appendix C, we further discuss about pixel rejection results and quantitative results according the number of sample.
\subsection{Discussion}
\textbf{Training time:} Our method increases training time due to the multiple predictions for uncertainty estimation. However, the sampling process is only executed at the second training step. Therefore, the total training time does not increase so much compare to training two ensemble models. Also, our method takes less time than training five ensemble models while achieving better performance.

\textbf{Generalization:} In this paper, we verify the effectiveness of our method by setting $M=2$. With only 2 submodels, experimental results show that our method can get plausible uncertainty and achieve comparable performance. Note that we can theoretically generate 2048 ensemble models with DenseNet ($M=2$ and $L=11$) and 262,144 ensemble models with U-Net ($M=2$ and $L=18$). Additional results with $M=3$ are attached to the Appendix F.

Deep ensembles have been shown to be a promising approach to not only uncertainty estimation but also increase the adversarial robustness \cite{lakshminarayanan2017simple,pang2019improving}. Therefore, it is interesting to generalize our idea to other tasks such as classification and adversarial attack detection. In the Appendix E, we briefly verify that our method is robust to adversarial attacks.

\textbf{Future work:} The generalization of our idea to different tasks and various network architectures would be an interesting research agenda. In the autonomous driving system, not only segmentation but also depth estimation is important. It would be meaningful to extend our idea to other datasets and various network structures. 

\begin{figure*}[t]
	\centering
	\includegraphics[width=0.77\textwidth]{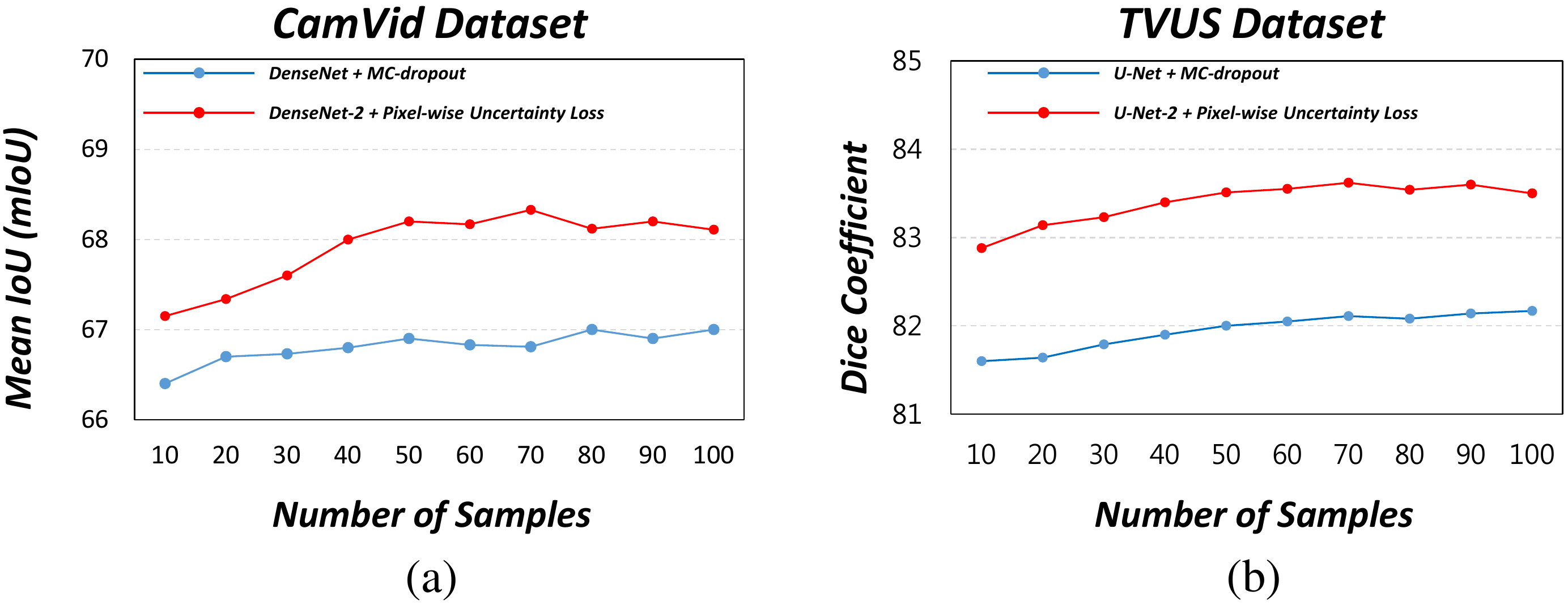}
    \vspace{-0.4cm}
	\caption{Effect of the number of sample on the performance. (a) Mean IOU measured on CamVid dataset. (b) Dice Coefficient measured on TVUS dataset.}
	\vspace{-0.2cm}
	\label{fig5}
\end{figure*} 

\section{Conclusion}
This paper presents an efficient ensemble model generation method for uncertainty estimation through Bayesian approximation in segmentation. To generate ensemble models with a few numbers of sub-models, the stochastic layer selection learning method is proposed. The stochastic layer selection learning methods enabled the efficient generation of ensemble models. Then, through the stochastic layer selection learning, our ensemble model can estimate uncertainty with Bayesian approximation. Furthermore, the pixel-wise uncertainty loss is devised to use the uncertainty of the model in training. The re-weighting of uncertain areas with large weights allows the model to improve on these samples. Comprehensive and comparative experiments show that the proposed method could provide useful uncertainty information which can be used in the quality control and improve the predictive performance.



\clearpage
%
%
\bibliographystyle{splncs04}
\bibliography{egbib}

\begin{thebibliography}{10}
\providecommand{\url}[1]{\texttt{#1}}
\providecommand{\urlprefix}{URL }
\providecommand{\doi}[1]{https://doi.org/#1}

\bibitem{amirul2017gated}
Amirul~Islam, M., Rochan, M., Bruce, N.D., Wang, Y.: Gated feedback refinement
  network for dense image labeling. In: Proceedings of the IEEE Conference on
  Computer Vision and Pattern Recognition. pp. 3751--3759 (2017)

\bibitem{ashukha2020pitfalls}
Ashukha, A., Lyzhov, A., Molchanov, D., Vetrov, D.: Pitfalls of in-domain
  uncertainty estimation and ensembling in deep learning. arXiv preprint
  arXiv:2002.06470  (2020)

\bibitem{badrinarayanan2017segnet}
Badrinarayanan, V., Kendall, A., Cipolla, R.: Segnet: A deep convolutional
  encoder-decoder architecture for image segmentation. IEEE transactions on
  pattern analysis and machine intelligence  \textbf{39}(12),  2481--2495
  (2017)

\bibitem{bhattacharyya2018long}
Bhattacharyya, A., Fritz, M., Schiele, B.: Long-term on-board prediction of
  people in traffic scenes under uncertainty. In: Proceedings of the IEEE
  Conference on Computer Vision and Pattern Recognition. pp. 4194--4202 (2018)

\bibitem{blundell2015weight}
Blundell, C., Cornebise, J., Kavukcuoglu, K., Wierstra, D.: Weight uncertainty
  in neural networks. arXiv preprint arXiv:1505.05424  (2015)

\bibitem{brostow2009semantic}
Brostow, G.J., Fauqueur, J., Cipolla, R.: Semantic object classes in video: A
  high-definition ground truth database. Pattern Recognition Letters
  \textbf{30}(2),  88--97 (2009)

\bibitem{chen2017deeplab}
Chen, L.C., Papandreou, G., Kokkinos, I., Murphy, K., Yuille, A.L.: Deeplab:
  Semantic image segmentation with deep convolutional nets, atrous convolution,
  and fully connected crfs. IEEE transactions on pattern analysis and machine
  intelligence  \textbf{40}(4),  834--848 (2017)

\bibitem{chen2018encoder}
Chen, L.C., Zhu, Y., Papandreou, G., Schroff, F., Adam, H.: Encoder-decoder
  with atrous separable convolution for semantic image segmentation. In:
  Proceedings of the European conference on computer vision (ECCV). pp.
  801--818 (2018)

\bibitem{dietterich2000ensemble}
Dietterich, T.G.: Ensemble methods in machine learning. In: International
  workshop on multiple classifier systems. pp. 1--15. Springer (2000)

\bibitem{efron1994introduction}
Efron, B., Tibshirani, R.J.: An introduction to the bootstrap. CRC press (1994)

\bibitem{feng2018towards}
Feng, D., Rosenbaum, L., Dietmayer, K.: Towards safe autonomous driving:
  Capture uncertainty in the deep neural network for lidar 3d vehicle
  detection. In: 2018 21st International Conference on Intelligent
  Transportation Systems (ITSC). pp. 3266--3273. IEEE (2018)

\bibitem{fort2019deep}
Fort, S., Hu, H., Lakshminarayanan, B.: Deep ensembles: A loss landscape
  perspective. arXiv preprint arXiv:1912.02757  (2019)

\bibitem{gal2016bayesian}
Gal, Y., Ghahramani, Z.: Bayesian convolutional neural networks with bernoulli
  approximate variational inference. International Conference on Learning
  Representations Workshop  (2016)

\bibitem{gal2016dropout}
Gal, Y., Ghahramani, Z.: Dropout as a bayesian approximation: Representing
  model uncertainty in deep learning. In: international conference on machine
  learning. pp. 1050--1059 (2016)

\bibitem{gal2017deep}
Gal, Y., Islam, R., Ghahramani, Z.: Deep bayesian active learning with image
  data. In: Proceedings of the 34th International Conference on Machine
  Learning-Volume 70. pp. 1183--1192. JMLR. org (2017)

\bibitem{gonen1989endometrial}
Gonen, Y., Casper, R.F., Jacobson, W., Blankier, J.: Endometrial thickness and
  growth during ovarian stimulation: a possible predictor of implantation in in
  vitro fertilization. Fertility and Sterility  \textbf{52}(3),  446--450
  (1989)

\bibitem{graves2011practical}
Graves, A.: Practical variational inference for neural networks. In: Advances
  in neural information processing systems. pp. 2348--2356 (2011)

\bibitem{gustafsson2019evaluating}
Gustafsson, F.K., Danelljan, M., Sch{\"o}n, T.B.: Evaluating scalable bayesian
  deep learning methods for robust computer vision. arXiv preprint
  arXiv:1906.01620  (2019)

\bibitem{hernandez2015probabilistic}
Hern{\'a}ndez-Lobato, J.M., Adams, R.: Probabilistic backpropagation for
  scalable learning of bayesian neural networks. In: International Conference
  on Machine Learning. pp. 1861--1869 (2015)

\bibitem{huang2018efficient}
Huang, P.Y., Hsu, W.T., Chiu, C.Y., Wu, T.F., Sun, M.: Efficient uncertainty
  estimation for semantic segmentation in videos. In: Proceedings of the
  European Conference on Computer Vision (ECCV). pp. 520--535 (2018)

\bibitem{jegou2017one}
J{\'e}gou, S., Drozdzal, M., Vazquez, D., Romero, A., Bengio, Y.: The one
  hundred layers tiramisu: Fully convolutional densenets for semantic
  segmentation. In: Proceedings of the IEEE conference on computer vision and
  pattern recognition workshops. pp. 11--19 (2017)

\bibitem{kendall2017bayesian}
Kendall, A., Badrinarayanan, V., Cipolla, R.: Bayesian segnet: Model
  uncertainty in deep convolutional encoder-decoder architectures for scene
  understanding. British Machine Vision Conference (BMVC)  (2017)

\bibitem{kendall2017uncertainties}
Kendall, A., Gal, Y.: What uncertainties do we need in bayesian deep learning
  for computer vision? In: Advances in neural information processing systems.
  pp. 5574--5584 (2017)

\bibitem{kendall2018multi}
Kendall, A., Gal, Y., Cipolla, R.: Multi-task learning using uncertainty to
  weigh losses for scene geometry and semantics. In: Proceedings of the IEEE
  conference on computer vision and pattern recognition. pp. 7482--7491 (2018)

\bibitem{kingma2014adam}
Kingma, D.P., Ba, J.: Adam: A method for stochastic optimization. International
  Conference on Learning Representations (ICLR)  (2014)

\bibitem{kirsch2019batchbald}
Kirsch, A., van Amersfoort, J., Gal, Y.: Batchbald: Efficient and diverse batch
  acquisition for deep bayesian active learning. In: Advances in Neural
  Information Processing Systems. pp. 7024--7035 (2019)

\bibitem{kundu2016feature}
Kundu, A., Vineet, V., Koltun, V.: Feature space optimization for semantic
  video segmentation. In: Proceedings of the IEEE Conference on Computer Vision
  and Pattern Recognition. pp. 3168--3175 (2016)

\bibitem{lakshminarayanan2017simple}
Lakshminarayanan, B., Pritzel, A., Blundell, C.: Simple and scalable predictive
  uncertainty estimation using deep ensembles. In: Advances in neural
  information processing systems. pp. 6402--6413 (2017)

\bibitem{li2019dfanet}
Li, H., Xiong, P., Fan, H., Sun, J.: Dfanet: Deep feature aggregation for
  real-time semantic segmentation. In: Proceedings of the IEEE Conference on
  Computer Vision and Pattern Recognition. pp. 9522--9531 (2019)

\bibitem{long2015fully}
Long, J., Shelhamer, E., Darrell, T.: Fully convolutional networks for semantic
  segmentation. In: Proceedings of the IEEE conference on computer vision and
  pattern recognition. pp. 3431--3440 (2015)

\bibitem{machtinger2005transvaginal}
Machtinger, R., Korach, J., Padoa, A., Fridman, E., Zolti, M., Segal, J.,
  Yefet, Y., Goldenberg, M., Ben-Baruch, G.: Transvaginal ultrasound and
  diagnostic hysteroscopy as a predictor of endometrial polyps: risk factors
  for premalignancy and malignancy. International Journal of Gynecologic Cancer
   \textbf{15}(2),  325--328 (2005)

\bibitem{minka2000bayesian}
Minka, T.P.: Bayesian model averaging is not model combination. Available
  electronically at http://www. stat. cmu. edu/minka/papers/bma. html pp.~1--2
  (2000)

\bibitem{nair2020exploring}
Nair, T., Precup, D., Arnold, D.L., Arbel, T.: Exploring uncertainty measures
  in deep networks for multiple sclerosis lesion detection and segmentation.
  Medical image analysis  \textbf{59},  101557 (2020)

\bibitem{osband2016deep}
Osband, I., Blundell, C., Pritzel, A., Van~Roy, B.: Deep exploration via
  bootstrapped dqn. In: Advances in neural information processing systems. pp.
  4026--4034 (2016)

\bibitem{pang2019improving}
Pang, T., Xu, K., Du, C., Chen, N., Zhu, J.: Improving adversarial robustness
  via promoting ensemble diversity. arXiv preprint arXiv:1901.08846  (2019)

\bibitem{park2019endometrium}
Park, H., Lee, H.J., Kim, H.G., Ro, Y.M., Shin, D., Lee, S.R., Kim, S.H., Kong,
  M.: Endometrium segmentation on tvus image using key-point discriminator.
  Medical physics  (2019)

\bibitem{postels2019sampling}
Postels, J., Ferroni, F., Coskun, H., Navab, N., Tombari, F.: Sampling-free
  epistemic uncertainty estimation using approximated variance propagation. In:
  Proceedings of the IEEE International Conference on Computer Vision. pp.
  2931--2940 (2019)

\bibitem{ronneberger2015u}
Ronneberger, O., Fischer, P., Brox, T.: U-net: Convolutional networks for
  biomedical image segmentation. In: International Conference on Medical image
  computing and computer-assisted intervention. pp. 234--241. Springer (2015)

\bibitem{roy2019bayesian}
Roy, A.G., Conjeti, S., Navab, N., Wachinger, C., Initiative, A.D.N., et~al.:
  Bayesian quicknat: model uncertainty in deep whole-brain segmentation for
  structure-wise quality control. NeuroImage  \textbf{195},  11--22 (2019)

\bibitem{rupprecht2017learning}
Rupprecht, C., Laina, I., DiPietro, R., Baust, M., Tombari, F., Navab, N.,
  Hager, G.D.: Learning in an uncertain world: Representing ambiguity through
  multiple hypotheses. In: Proceedings of the IEEE International Conference on
  Computer Vision. pp. 3591--3600 (2017)

\bibitem{seo2019learning}
Seo, S., Seo, P.H., Han, B.: Learning for single-shot confidence calibration in
  deep neural networks through stochastic inferences. In: Proceedings of the
  IEEE Conference on Computer Vision and Pattern Recognition. pp. 9030--9038
  (2019)

\bibitem{snoek2019can}
Snoek, J., Ovadia, Y., Fertig, E., Lakshminarayanan, B., Nowozin, S., Sculley,
  D., Dillon, J., Ren, J., Nado, Z.: Can you trust your model's uncertainty?
  evaluating predictive uncertainty under dataset shift. In: Advances in Neural
  Information Processing Systems. pp. 13969--13980 (2019)

\bibitem{srivastava2014dropout}
Srivastava, N., Hinton, G., Krizhevsky, A., Sutskever, I., Salakhutdinov, R.:
  Dropout: a simple way to prevent neural networks from overfitting. The
  journal of machine learning research  \textbf{15}(1),  1929--1958 (2014)

\bibitem{tieleman2012lecture}
Tieleman, T., Hinton, G.: Lecture 6.5-rmsprop: Divide the gradient by a running
  average of its recent magnitude. COURSERA: Neural networks for machine
  learning  \textbf{4}(2),  26--31 (2012)

\bibitem{tramer2017ensemble}
Tram{\`e}r, F., Kurakin, A., Papernot, N., Goodfellow, I., Boneh, D., McDaniel,
  P.: Ensemble adversarial training: Attacks and defenses. International
  Conference on Learning Representations (ICLR)  (2018)

\bibitem{valdenegro2019deep}
Valdenegro-Toro, M.: Deep sub-ensembles for fast uncertainty estimation in
  image classification. arXiv preprint arXiv:1910.08168  (2019)

\bibitem{welling2011bayesian}
Welling, M., Teh, Y.W.: Bayesian learning via stochastic gradient langevin
  dynamics. In: Proceedings of the 28th international conference on machine
  learning (ICML-11). pp. 681--688 (2011)

\bibitem{yu2018bisenet}
Yu, C., Wang, J., Peng, C., Gao, C., Yu, G., Sang, N.: Bisenet: Bilateral
  segmentation network for real-time semantic segmentation. In: Proceedings of
  the European Conference on Computer Vision (ECCV). pp. 325--341 (2018)

\bibitem{yu2015multi}
Yu, F., Koltun, V.: Multi-scale context aggregation by dilated convolutions.
  arXiv preprint arXiv:1511.07122  (2015)

\bibitem{yuan2019object}
Yuan, Y., Chen, X., Wang, J.: Object-contextual representations for semantic
  segmentation. arXiv preprint arXiv:1909.11065  (2019)

\bibitem{zhao2017pyramid}
Zhao, H., Shi, J., Qi, X., Wang, X., Jia, J.: Pyramid scene parsing network.
  In: Proceedings of the IEEE conference on computer vision and pattern
  recognition. pp. 2881--2890 (2017)

\bibitem{zhu2019improving}
Zhu, Y., Sapra, K., Reda, F.A., Shih, K.J., Newsam, S., Tao, A., Catanzaro, B.:
  Improving semantic segmentation via video propagation and label relaxation.
  In: Proceedings of the IEEE Conference on Computer Vision and Pattern
  Recognition. pp. 8856--8865 (2019)

\end{thebibliography}


\section*{Supplementary material}

# Appendix A. Training and Testing Algorithm

In this section, we describe in detail the proposed training and testing schemes. As described in the paper, our proposed training scheme consists of two stages. In the first step, every iteration, we sample the layer with $P_l(m)$ and sample a network. The weight of the sampled network at $i^{th}$ iteration can be defined as $w_i^{Sample}$. Then, we update the weight of the sampled network. We continue this process until the validation loss is saturated ($t$). Through the first stage, the network could predict reliable segmentation results and estimate uncertainty with stochastic layer selection learning. In the second stage, we leverage the estimated uncertainty map. Then, the model could gradually learn from the uncertain pixels which are easier than hard negative cases in training. To estimate the uncertainty map, we predict sample outputs from the sampled network. We sample $N$ prediction $S=\{\hat{S}_1,\hat{S}_2,...,\hat{S}_N\}$ and estimate uncertainty with the pixel-wise variance of samples $U=var\{\hat{S}_1,\hat{S}_2,...,\hat{S}_N\}$. Then, we update sample network with weight parameter is $w_j^{Sample}$ where $w$' and $j$ denote the weight of the sampled network and iteration of the second stage, respectively. The network optimized by $\mathcal{L}_{total}$.

In the test step, we sample $N$ prediction with $w_n^{Sample}$. Then, we predict final segmentation output by mean of the sample outputs. Following is the algorithm of the proposed learning and testing scheme.

---

**Algorithm 1:** Pseudocode of the training and the test procedure for our method

---

**Training:** For given training data $X$, training label $Y$, number of sample $N$, and total training iteration $T$

$1^{st}$ **step**

**for** $i = 1, 2, \cdots, t$ **do**
  Select mini-batch set $(x_i, y_i) \sim (X, Y)$ in training dataset
  Generate sample network with $P_l$,
  where $P_l \sim categorical(x_1, x_2, ..., x_M; \mu_1, \mu_2, ..., \mu_M)$
  Minimize $\mathcal{L}_{var}(x_i, y_i, w_i^{Sample})$, where $w_i^{Sample}$ is the weight of sample network
**end**

$2^{nd}$ **step**

**for** $j = t+1, \cdots, T$ **do**
  Select mini-batch set $(x_i, y_i)$ in training dataset
  **for** $n = 1, \cdots, N$ **do**
    Randomly sample prediction $S_n$ with $w_n^{Sample}$
  **end**
  Estimate uncertainty with sample prediction
  $\mathbf{S} = \{\hat{S}_1, \hat{S}_2, ..., \hat{S}_N\}$
  $\mathbf{U} = var\{\hat{S}_1, \hat{S}_2, ..., \hat{S}_N\}$
  Normalize the $U \sim (0, 1)$
  Minimize $\mathcal{L}_{total}(x_j, y_j, w_j^{Sample}, U)$
**end**

---

**Test:** For given test image $\mathbf{x}$, sample mean output $\mathbf{y}$, and uncertainty estimation $\mathbf{u}$

**for** $n = 1, \cdots, N$ **do**
  Randomly sample prediction $S_n$ with $w_n^{Sample}$
**end**

Final prediction $\mathbf{y}=mean(\hat{S}_1, \hat{S}_2, ..., \hat{S}_N)$

Estimated uncertainty $\mathbf{u}=var\{\hat{S}_1, \hat{S}_2, ..., \hat{S}_N\}$

---

## Appendix B. More Qualitative Results

In this section, we show more qualitative results on CamVid dataset. Figure 6 shows the qualitative results of our proposed method. The yellow circle points out the misclassified region. However, the model estimate there as an uncertain region. Therefore, we can use the estimated uncertainty map for quality control.

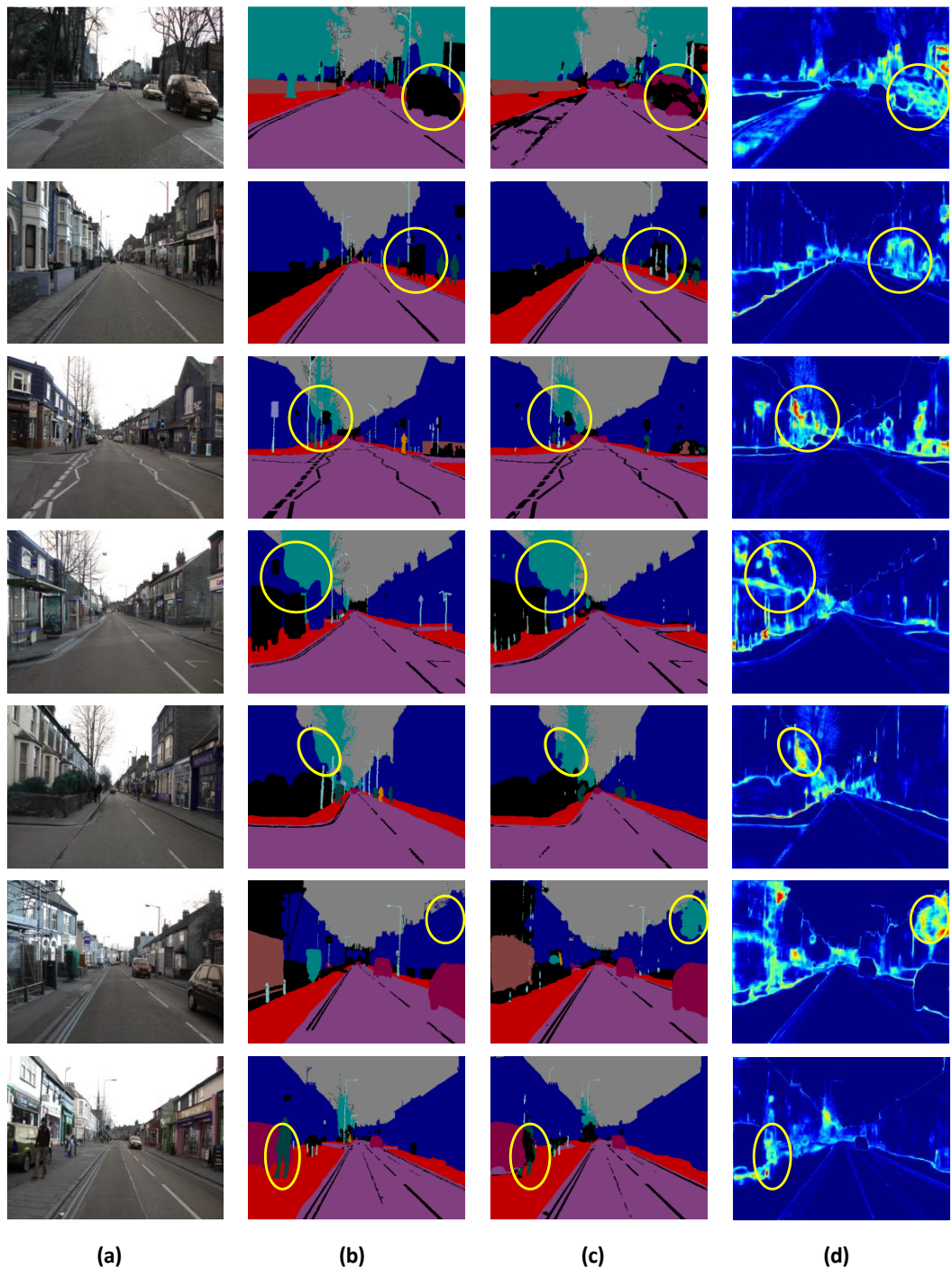

**Fig. 6.** Qualitative results of segmentation and uncertainty estimation in our method on CamVid dataset. (a) input images, (b) ground truth segmentation maps, (c) our prediction results with 50 samples, and (d) an uncertainty estimation results. The yellow circle denotes incorrect segmentation regions in (c) while highly uncertain is estimated in (d).

# Appendix C. Quantitative Evaluation with Uncertainty-based Pixel Rejection

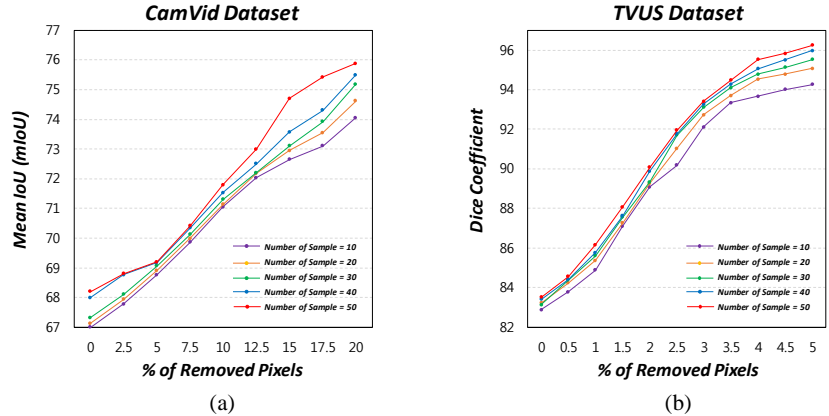

(a)

(b)

**Fig. 7.** Quantitative evaluation with uncertainty-based pixel rejection according to the number of samples. We sample the prediction from 10 to 50. (a) shows mIoU according to pixel-wise rejection on Cavid dataset. (b) shows Dice coefficient according to pixel-wise rejection on TVUS dataset.

In this section, we show the uncertainty-based pixel rejection results according to the number of samples. In the experiment, we get the mean prediction and uncertainty estimation results from 10 to 50 samples. Figure 7 (a) shows the result on CamVid dataset and (b) shows the result on TVUS dataset. As shown in the figure, as the removed pixels are increased, the performance also increases. The slopes of the methods with different numbers of samples are similar, which implies that our method could efficiently estimate the uncertainty.

# Appendix D. Qualitative Comparison According to the Number of Sampling

In this section, we compare the semantic segmentation results according to the number of samples. In this experiment, we evaluate DenseNet + MC-dropout and DenseNet-2+Pixel-wise Uncertainty Loss (Ours) method. Figure 8 shows the mean prediction results on CamVid dataset. As shown in the figure, our method predicts better mean prediction with only 10 samples compared with the MC-dropout method which uses 50 samples. It is also verified.

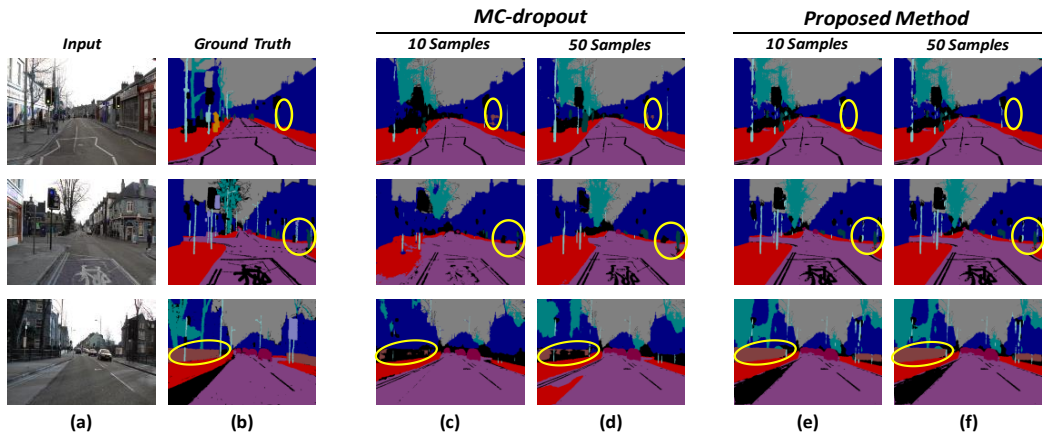

**Fig. 8.** Mean prediction results according to the number of samples. (a) the input images, (b) ground truth segmentation maps, (c) and (d) are the mean prediction of the DenseNet + MC-dropout method with 10 and 50 samples, respectively. (e) and (f) are the mean prediction of the DenseNet-2 + Pixel-wise Uncertainty Loss (ours) with 10 and 50 samples, respectively.

# Appendix E. Robustness against Adversarial Attacks

As discussed in the manuscript, the deep ensembles can be used in various fields such as uncertainty estimation and improve the robustness against to adversarial example. It is also known that deep ensemble models are more robust than MC-drop based approaches at the same number of sampling [18, 42]. Therefore, in this section, we verify the robustness of our proposed method against to adversarial example. To verify this, we generate adversarial examples by an FGSM algorithm [R1] with epsilon=0.003 on CamVid dataset. Then, we investigate how many pixels should be rejected to recover the original performance. For the rejection experiments, the pixels with high uncertainty pixels are rejected in descending order. To this end, we normalize the estimated uncertainty map from 0 to 1. By rejecting uncertain pixels from 0% to 20% with an interval of 2.5% in the test adversarial examples, we measure the mIoU. Table 4 shows the pixel-wise rejection results. As shown in Table 4, our method could achieve the original performance (mIoU of 68.1%) by rejecting only 10% of pixels. However, in the case of DenseNet + MC-dropout, it should reject about 15% pixels to recover original performance (66.6%). Also, our proposed method shows better performance than DenseNet + MC-dropout over all rejection rates. Since our proposed method selects layers randomly, it is hard to get gradients. Therefore, our proposed method could be robust to adversarial examples.

Table 4. Uncertainty based pixel rejection result on CamVid dataset. The test images are corrupted by the FGSM algorithm. mIoU is measured over the different pixel rejection rates.

| | % of pixel rejection rate | | | | | | | | |
|---|---|---|---|---|---|---|---|---|---|
| | 0 | 2.5 | 5 | 7.5 | 10 | 12.5 | 15 | 17.5 | 20 |
| DenseNet + MC-dropout | 60.20 | 60.94 | 61.95 | 63.09 | 64.24 | 65.43 | **66.56** | 67.68 | 69.71 |
| DenseNet-2+Pixel-wise Uncertainty Loss (Ours) | 62.64 | 63.85 | 65.00 | 66.00 | **68.10** | 69.15 | 69.86 | 69.99 | 70.64 |

# Appendix F. Generalization Experiment

Table 5. Mean IoU comparison with the ensemble models and our proposed method with $M$=2 and $M$=3. The experiments have been conducted on CamVid dataset.

|  | mIou | # of Parameters (M) |
|---|---|---|
| Ensemble (1) | 66.4 | 9.4 |
| Ensemble (2) | 66.8 | 18.8 |
| Ensemble (3) | 67.0 | 28.2 |
| Ensemble (4) | 67.1 | 37.6 |
| Ensemble (5) | 67.2 | 47.0 |
| DenseNet-2 | 67.4 | 18.8 |
| DenseNet-3 | 68.1 | 28.2 |

Table 6. Dice Coefficient comparison with ensemble model and our proposed method with $M$=2 and $M$=3. The experiment conduct on TVUS dataset.

|  | Dice Coefficient | # of Parameters (M) |
|---|---|---|
| Ensemble (1) | 82.30 | 31.0 |
| Ensemble (2) | 82.51 | 62.0 |
| Ensemble (3) | 83.12 | 93.1 |
| Ensemble (4) | 83.20 | 124.1 |
| Ensemble (5) | 83.38 | 155.1 |
| U-Net-2 | 83.51 | 62.0 |
| U-Net-3 | 84.01 | 93.1 |

**Quantitative Results on $M$=3**

In this section, we verify the effect of $M$=3. Table 5 and Table 6 show quantitative comparison with basic ensemble models and our proposed method. In the experiment, we also use 50 sample predictions. As shown in Table 5, DenseNet-3 shows better performance than Ensemble (5) while having less parameters. Compared to Ensemble (3) which has the same number of parameter, our proposed method shows 1.1% higher performance. Table 6 shows the Dice Coefficient score with U-Net-3 and TVUS dataset. Also in the case of U-Net-3, it shows better performance than Ensemble (5) while having less parameter. It shows 0.99% higher performance than Ensemble (3). These results show that the proposed method could be also generalized to the case $M$=3. The proposed method could effectively generate the ensemble models with small number of subnetwork.

**Generalization of the Pixel-wise Uncertainty Loss**

Table 7. Comparison of DenseNet + MC-dropout with different training stragegies.

| Method | mIoU |
| --- | --- |
| DenseNet + MC-dropout | 66.4 |
| DenseNet + MC-dropout + Pixel-wise uncertainty Loss | 67.2 |

The concept of the proposed pixel-wise uncertainty loss it to encourage the uncertain region, then gradually overcome its limitation from uncertain pixels. Therefore, our proposed pixel-wise uncertainty loss could be generalized to MC-dropout based segmentation method. To verify this, we conducted an experiment with DenseNet+MC-dropout method and CamVid dataset. In the experiment, we trained DenseNet with dropout layers with cross-entropy loss. After the network saturated, we fine-tuned the network with pixel-wise uncertainty loss. The experiment result shows in Table 7. As shown in the table, our proposed method improves the performance by overcoming the uncertain pixel. It improves 0.8% compared to DenseNet + MC-dropout method. The results verify that our proposed pixel-wise uncertainty loss can be generalized to other uncertainty estimation methods.

# Reference

[R1] Goodfellow, Ian J., Jonathon Shlens, and Christian Szegedy. "Explaining and harnessing adversarial examples." *arXiv preprint arXiv:1412.6572* (2014).


\end{document}